\documentclass[runningheads]{llncs}
\usepackage{graphicx}
\usepackage{cite}
\usepackage{amsmath,amssymb,amsfonts}
\usepackage{textcomp}
\usepackage{xcolor}
\usepackage{listings}
\usepackage{amsmath,amssymb,exscale}
\usepackage{blindtext, graphicx}
\usepackage{verbatim}
\usepackage{float}
\usepackage{algpseudocode}
\usepackage{algorithm}
\usepackage{fancyvrb}
\usepackage{bera}
\usepackage{mathtools}
\usepackage{lipsum}
\usepackage{enumitem}
\usepackage{subcaption}
\usepackage[noend]{algcompatible}
\usepackage[hyphenbreaks]{breakurl}
\usepackage[hyphens]{url}

\newcommand{\mc}[1]{\ensuremath{\mathcal{#1}}}
\newcommand{\mbf}[1]{\ensuremath{\mathbf{#1}}}
\newcommand{\st}{such that}

\algdef{SE}[DOWHILE]{Do}{doWhile}{\algorithmicdo}[1]{\algorithmicwhile\ #1}%

\begin{document}

\title{On Understanding the Influence of Controllable Factors
 with a Feature Attribution Algorithm:\\ a Medical Case Study}
\titlerunning{Controllable Factor Feature Attribution}
%
\author{Veera Raghava Reddy Kovvuri\inst{1} \and
Siyuan Liu\inst{2} \and
Monika Seisenberger\inst{1} \and\\
Berndt M\"{u}ller\inst{1}\and
Xiuyi Fan\inst{2}}
\authorrunning{V.R.R. Kovvuri, S. Liu, M. Seisenberger, B. M\"{u}ller, X. Fan}

%
\institute{Swansea University, United Kingdom \and
Nanyang Technological University, Singapore}
\maketitle{}             

\begin{abstract}
Feature attribution XAI algorithms enable their users to gain insight into the 
underlying patterns of large datasets through their feature importance calculation. Existing feature attribution algorithms treat all features in a dataset homogeneously, which may lead to misinterpretation of consequences of changing feature values. In this work, we consider partitioning features into {\em controllable} and {\em uncontrollable} parts and propose the \emph{Controllable fActor Feature Attribution} (CAFA) approach to compute the relative importance of controllable features. We carried out experiments applying CAFA to two existing datasets and our own COVID-19 non-pharmaceutical control measures dataset. Experimental results show that with CAFA, we are able to exclude influences from uncontrollable features in our explanation while keeping the full dataset for prediction. 

\end{abstract}

\section{Introduction}

Feature attribution algorithms \cite{Lundberg2017} are a popular class
of Explainable AI (XAI) algorithms. Given a prediction instance, they tell the
relative ``importance'' of each feature in the instance. In addition
to ``explaining'' the prediction model, importance measures also
reveal insight about the instance being explained - \cite{Antoniadi2021} show 
that XAI can help "generating the hypothesis about causality" in developing 
decision support systems. In this sense,
feature attribution algorithms are seen as a data mining tool for
extracting and discovering patterns in large datasets. For instance,
\cite{Duell21} use feature attribution algorithms to understand
important factors affecting cancer patient survivability; 
\cite{Fan20b} use feature attribution algorithms to study
factors affecting the transmission of SARS-CoV-2; and \cite{Liu21} use feature attribution to analyse factors affecting foreign exchange markets.

However, existing feature attribution algorithms (see e.g.,
\cite{Tjoa2020,Adadi2018,molnar2019} for overviews) treat all
features homogeneously when computing their relative importances. Such
homogeneity may not always give desirable interpretations when
feature attribution algorithms are used for data mining purposes.
Consider the following hypothetical example.

\begin{quote}
Suppose we want to estimate the chance for some individual having
breast cancer, with features like {\em age}, {\em gender}, {\em
  weight}, {\em alcohol intake}, {\em smoking habits}, {\em family history}, etc. A
predictive model estimates the likelihood of the person having breast
cancer; and a feature attribution algorithm gives attributions like
{\em age: 0.3}, {\em gender: 0.13}, {\em weight: 0.27}, {\em alcohol
  intake: 0.15}, {\em smoking: 0.3}, {\em family history: 0.36}, etc.
  \vspace{-5pt}
\end{quote}

From these calculated values, we notice that certain features, such as {\em
  age}, {\em gender} and {\em family history}, while being influential
to the prediction, are {\em uncontrollable risk factors} \cite{Draper06}.
Knowing the relative importance of these features makes little
contribution to clinical decision making. On the other hand, features
representing {\em controllable risk factors} such as {\em weight},
{\em alcohol intake} and {\em smoking habits} are vital to clinical
interventions \cite{Draper06}. Thus, from an intervention
perspective, it is necessary to distinguish these two classes of
factors and compute their influences accordingly. We raise the question:

\centerline{
{\em What are the influences of controllable factors used in a prediction?}
}

To answer this question, a naive approach would be to build another predictive
model, which only considers controllable factors, and apply
feature attribution algorithms to that model.  However, as explained
in \cite{IndreZ2016}, dropping features from models can negatively impact the
model performance. Thus, instead of building models with
fewer features, we suggest creating algorithms that
treat controllable factors differently from uncontrollable ones.

In this paper, we present {\em Controllable fActor Feature Attribution
  (CAFA)}. Through {\em selective perturbation} and {\em
  global-for-local interpretation}, CAFA
computes the relative importance of controllable factors for individual
instances using prediction models built from all features. We
apply CAFA on lung cancer data in Simulacrum\footnote{Simulacrum is a dataset developed by Health Data Insight CiC derived from
anonymous cancer data provided by the National Cancer Registration
and Analysis Service, which is part of Public Health England.} and UCI breast cancer dataset\footnote{ \url{http://archive.ics.uci.edu/ml/datasets/Breast+Cancer}} to study the influence of controllable factors on survival time or recurrence. In a second experiment, we apply CAFA to a COVID-19 virus
transmission case study, for identifying the effectiveness of
non-pharmaceutical control measures.

\vspace{-10pt}

\section{Background}
\label{sec:bkg}
\vspace{-5pt}

Given a prediction model $f \in \mc{F}$ where $\mc{F}$ is a set of
models, let $\mbf{y} = f(\mbf{x})$ be the prediction
made by $f$ on the input $\mbf{x} = \langle \mbf{x}_1, \ldots, \mbf{x}_m
\rangle \in \mathbb{R}^m$, a {\em feature attribution} algorithm give
an explanation $\Phi_\mbf{x} = \langle \phi_1, \ldots, \phi_m \rangle \in
\mathbb{R}^m$, where $\phi_i$ can
be viewed as the relative importance of $\mbf{x}_i$ for $\mbf{y} = f(\mbf{x})$.
We briefly review the two algorithms supporting this work as follows.

{\bf Local interpretable model-agnostic explanations (LIME)}
\cite{Ribeiro16}. To explain how a model $f$ predicts a data
instance $\mbf{x}$, LIME generates a new dataset $D = \{(\mbf{x}_1,
f(\mbf{x}_1)), \ldots, (\mbf{x}_n,f(\mbf{x}_n))\}$ consisting of
$n$ perturbed samples $\mbf{x}_1,\ldots,\mbf{x}_n$ within some proximity
$\pi_\mbf{x}$ of $\mbf{x}$, and then fits a an interpretable model
$g$ with $D$. Parameters of the new model are the explanation of
$\mbf{x}$.  Formally, LIME computes explanations as:
\begin{equation}
  \text{LIME}(\mbf{x}) = \arg\min_{g \in \mc{G}} L(f,g,\pi_{\mbf{x}})+\Omega(g),
\end{equation}
where $L$ is a loss function comparing $f$ and $g$, $\mc{G}
\subseteq \mc{F}$ is a class of interpretable models, and ${\Omega(g)}$
the complexity of $g$.

{\bf SHapley Additive exPlanations (SHAP)} \cite{Lundberg2017}
is based on the coalitional game theory concept of a {\em Shapley value},
assigned to each feature of instance $\mbf{x}$. The  Shapley value of
a feature is its marginal contribution to the prediction thus explains
the prediction. Specifically, let $g$ be the explanation model. For
an instance $\mbf{x}$ with $m$ features, there is a corresponding $z \in
\{0,1\}^m$ \st{} SHAP specifies $g$ being a linear function of $z$:
\begin{equation}
  g(z) = \phi_0 + \sum_{j=1}^{m} \phi_j z_{j},
\end{equation}
where $\phi_j (0 < j \leq m)$ is the Shapley value of feature $j$ and 
$\phi_0$ is the ``average'' prediction when none of the features in $\mbf{x}$ is present.

Both SHAP and LIME are local methods in the sense that they explain 
individual instances in a dataset. Global explanation, which describes 
the average behaviour of the dataset, can be simply obtained by taking 
the average of local explanations of instances in the dataset \cite{molnar2019}.

\vspace{-10pt}
\section{Our approach}
\label{sec:approach}
\vspace{-5pt}

\begin{figure}[!ht]
  \centerline{
    \includegraphics[width=1\textwidth,trim={0cm 0.3cm 0cm 0.3cm},clip]{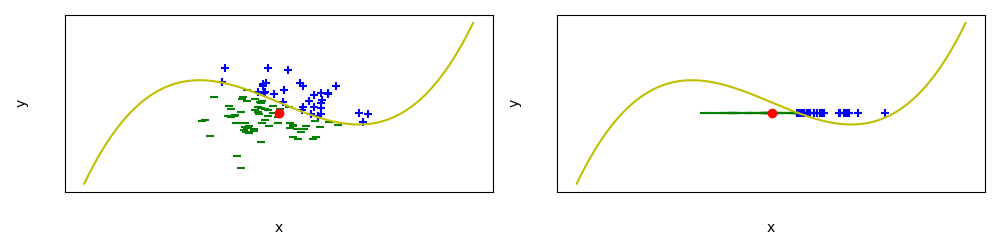}
    }
    \vspace{-10pt}
  \caption{Selective Perturbation in CAFA. The point of interest (explanation point) and the generated dataset are shown in the figures. The red dot denotes the point of interest in a 2D space. The yellow curve is the decision boundary. Blue ``+'' and green ``-'' denote generated positive and negative samples, respectively. The figure on the left illustrates the standard perturbation (LIME), where both features $x$ and $y$ are perturbed; the figure on the right illustrates the selective perturbation (CAFA), where only the $x$ axis, representing the controllable factor, is perturbed. \label{fig:pert}
  }
\vspace{-5pt}
\end{figure}

CAFA computes feature importances for controllable factors through {\em
  selective perturbation} and {\em global-for-local
  interpretation}. Concepturally, CAFA is inspired by LIME \st{} a set
of perturbed samples is generated to compute the feature
importance. However, there are two main differences. Firstly,
unlike LIME where the perturbation is carried out uniformly
throughout all features, CAFA selectively perturbs features
representing controllable factors. Secondly, with the dataset
generated, instead of fitting a weak interpretable model for
computing explanations, a strong model is chosen to fit the dataset.
We then determine the feature importance of controllable factors by
using an explainer to compute the global explanation on the 
dataset. Fig.~\ref{fig:pert} illustrates CAFA's selective
perturbation strategy.

Given a prediction model $f$, for a data point $\mbf{x}$ with $m$
features partitioned into two sets $F_c$ ({\em controllable})
and $F_u$ ({\em uncontrollable}) \st{} $F_c \cap F_u =
\{\}$, to compute feature importance for $F_c$, we construct a data
set with $n$ points
$$D_{\mbf{x}} = \{(\mbf{x}_1, f(\mbf{x}_1)), \ldots, (\mbf{x}_n,
f(\mbf{x}_n))\}$$ \st{} for all $(\mbf{x}_i, f(\mbf{x}_i)) \in
D_{\mbf{x}}$, the following two conditions hold:

\begin{enumerate}[nolistsep]
\item
  $\delta(\mbf{x}, \mbf{x}_i) \leq \pi_{\mbf{x}}$, where $\delta$ is a
  distance function and $\pi_{\mbf{x}}$ is some proximity threshold, and

\item
  for $\mbf{x} = \langle v_1, \ldots, v_m \rangle$, and $\mbf{x}_i =
  \langle v_1^i, \ldots, v_m^i \rangle$, for all $j  (1 \leq j \leq
  m)$, it is the case that if feature $j$ is in $F_u$, then $v_j
  = v_j^i$.
\end{enumerate}
For two instances $\mbf{x}_1 = \langle v^1_1, \ldots, v^1_m \rangle$
and $\mbf{x}_2 = \langle v^2_1, \ldots, v^2_m \rangle$, the distance
function $\delta(\mbf{x}_1, \mbf{x}_2)$ is
\begin{equation}
\delta(\mbf{x}_1, \mbf{x}_2) = \frac{\sum_{i=1}^m \omega_i d(v^1_i, v^2_i)}{\sum_{i=1}^m \omega_i} ,
\end{equation}
where $\omega_i$ is the weight of feature $i$ and $d(v^1_i, v^2_i)$ is
defined by\footnote{Note that we assume some standard normalization / scaling pre-processing is performed on the dataset so all continuous features take values in the range [0,1].}, 
\begin{itemize}
\item
if feature $i$ is categorical, then
\begin{equation}
d(v^1_i, v^2_i) =
  \begin{cases}
    0 & \quad \text{if } v^1_i = v^2_i \text{,} \\
    1  & \quad \text{otherwise;}
  \end{cases}
\end{equation}

\item
  if feature $i$ is continuous, then
\begin{equation}
  d(v^1_i, v^2_i) = |v^1_i -v^2_i|.
\end{equation}
\end{itemize}

We then build a strong prediction model $g$ from $D_\mbf{x}$ and calculate the global explanation $g(D_\mbf{x})$ using SHAP by first computing local explanations for all instances in $D_\mbf{x}$ and then averaging the results. Overall, for an instance $\mbf{x}$ and explanations $\Phi_i$ computed over $D_\mbf{x}$,

\begin{equation}
  \text{CAFA}(\mbf{x}) = \frac{1}{n}\sum_{i=1}^n\Phi_i.
\end{equation}
Thus, we use the {\em global} explanation computed with a strong predictor on $D_\mbf{x}$ as the {\em local} explanation for $\mbf{x}$. This {\em global-for-local interpretation} is superior to LIME's local surrogate approach, 
as it has been shown that SHAP is more robust than LIME \cite{Slack20,Honegger18,Lundberg19,Yalcin21}.

Algorithm~\ref{alg:CAFA} describes the process in detail. Since all points in $D_\mbf{x}$ have same values for their uncontrollable features, these features have no correlation to class labels of points in $D_\mbf{x}$. Thus, their feature importance will be assigned to 0, as they make no contribution to the prediction.
By setting that each class contains $K$ samples (Line~\ref{line:K}), we ensure that $D_\mbf{x}$ is balanced.

\vspace{-15pt}
\begin{algorithm}
  \caption{Selective Perturbation and Global-for-Local Interpretation.\label{alg:CAFA}}

{\bf Input:} Data point $\mbf{x}$, Prediction model $f$, Proximity threshold $\pi_{\mbf{x}}$, Distance Function $\delta$, Controllable features $F_c$, Sample class size $K$

{\bf Output:} Feature Importance $\Phi$

\begin{algorithmic}[1]

\STATE Let $D_\mbf{x}' = []$;

\Do
  \STATE Randomly generate a data point $\mbf{x}'$ such that for all features $v \in F_{u}$, $\mbf{x}'$ contains the same value as $\mbf{x}$ in $v$ and
  $\delta(\mbf{x},\mbf{x}') \leq \pi_\mbf{x}$;
  \STATE Append $(\mbf{x}',f(\mbf{x}'))$ to $D_\mbf{x}'$;
  \STATE Let $r$ be the size of the smallest class in $D_\mbf{x}'$;
\doWhile{$r < K$};

\STATE Construct $D_\mbf{x}$ from $D_\mbf{x}'$ by sampling $K$ elements from each class in $D_\mbf{x}'$; \label{line:K}

\STATE Let $\Phi$ be the global explanation for $g(D_\mbf{x})$ with a strong predictor $g$;

\STATE \textbf{return} $\Phi$;
\label{algoritham1}
\end{algorithmic}
\end{algorithm}
\vspace{-35pt}


\section{Experiments with Two Existing Medical Datasets}
\label{sec:exp}

\vspace{-10pt}

As an experiment, we apply CAFA to the lung cancer data in Simulacrum and the UCI breast cancer dataset. We predict 12-months survival on the lung cancer dataset, which contains 2,242 instances specified by 28 features:
\begin{itemize}[nolistsep]
  \item Four uncontrollable features: age, ethnicity, sex and height;
  \item 20 controllable features: morph, weight, dose administration, regimen outcome description, administration route, clinical trial, cycle number, regimen time delay, cancer plan, T best, N best, grade, CReg code, laterality, ACE, CNS, performance, chemo radiation, regimen stopped early, 
  and M Best.\footnote{Description of features used in this dataset can be found at the Cancer Registration Data Dictionary and the SACT Data Dictionary, with links available at:\\ \url{https://simulacrum.healthdatainsight.org.uk/available-data/table-descriptions/}.}
\end{itemize}

The breast cancer dataset comprises 286 data instances, predicting cancer recurrence, each containing 9 features, which are:
\begin{itemize}[nolistsep]
  \item Two uncontrollable features: age and menopause;
  \item Seven numerical controllable features: tumor size, inv-nodes, node-caps, deg-malig, breast, breast-quad, and irradiate. 
\end{itemize}
Random forest classifiers are used in both cases.

\begin{figure}[!ht]
     \begin{subfigure}[a]{\linewidth}
         \centering
         \includegraphics[width=1\linewidth]{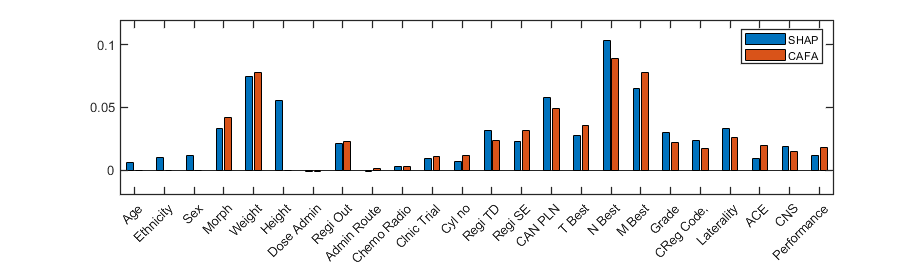}
         \caption{A lung cancer instance randomly selected from the Simulacrum dataset. Uncontrollable features are: {\em Age}, {\em Ethnicity}, {\em Sex}, and {\em Height}.}
         \label{fig2a}
     \end{subfigure}
     \begin{subfigure}[b]{\linewidth}
         \centering
         \includegraphics[width=1\linewidth]{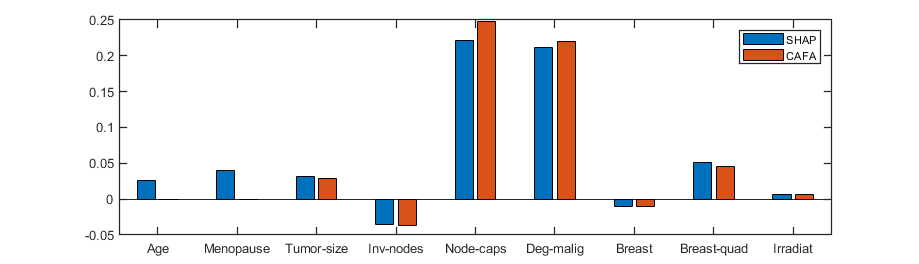}
         \caption{A breast cancer instance randomly selected from the UCI breast cancer dataset. Uncontrollable features are {\em Age} and {\em Menopause}.}
         \label{fig2b}
     \end{subfigure}
     
   \caption{Illustration of CAFA vs. SHAP on two explanation instances selected from two medical datasets. We observe that (1) with CAFA, all uncontrollable features are assigned importance 0; (2) for controllable features, CAFA produces results that are agreeable with the ones given by SHAP.\label{fig:lung_cancer_local}}
   \vspace{-10pt}
\end{figure}

Firstly, we illustrate the influence of controllable features on prediction results on individual instances (local explanations). To this end, we randomly sample an instance from each dataset, as follows:
\begin{itemize}[nolistsep]
\item
{\em Lung Cancer: age 71; ethnicity 5; sex 0; morph 8140; weight 49.8; height 1.83; dose administration 8; regimen outcome 1; administration route 1; clinical trial 2; cycle number 1; regimen time delay 0; cancer plan 0; T Best 3; N Best 0; grade 3; CReg Code 401; laterality 2; ACE 9; CNS 99; performance 0; chemo radiation 0; regimen stopped early 1; M Best 0.}
 
\item
{\em Breast Cancer: age 40;  menopause 0;  tumor-size 6;  inv-nodes 0; node-caps 1; deg-malig 3;  breast 0; breast-quad 3; irradiate 0.}
\end{itemize}
For each instance $\mbf{x}$, we generate $D_\mbf{x}$ containing 1,000 
perturbed instances (binary classification, $K=500$) 
and carry out CAFA calculation as shown in Algorithm~\ref{alg:CAFA}. We let $\pi_\mbf{x}$ be the average distance between points and feature weights $\omega_i = 1$.
Results from SHAP and CAFA
are shown in Fig.~\ref{fig:lung_cancer_local}. In this figure, the x-axis shows the features; y-axis shows feature importance. For each feature, the left (blue) bar shows the SHAP result of the feature, and the right (red) bar shows the importance calculated with CAFA. We observe that:
\begin{enumerate}[nolistsep]
\item
For uncontrollable features, i.e., ``age'', ``ethnicity'', ``sex'', and ``height'' from the lung cancer dataset as well as ``age'' and ``menopause'' from the breast cancer dataset, 
the assigned importance value is 0, as expected;

\item
For controllable features, there is a strong correlation, 0.96 for lung cancer and 0.99 for breast cancer, between values represented by the blue and the red bars, suggesting that CAFA is agreeable with SHAP.
\end{enumerate}
This suggests that CAFA successfully excludes influences of uncontrollable features with its calculation, while maintaining properties of standard feature attribution algorithms such as SHAP. 

\begin{figure}[!ht]
     \begin{subfigure}[a]{\linewidth}
         \centering
         \includegraphics[width=0.9\linewidth,trim={0cm 0.7cm 0cm 0.0cm},clip]{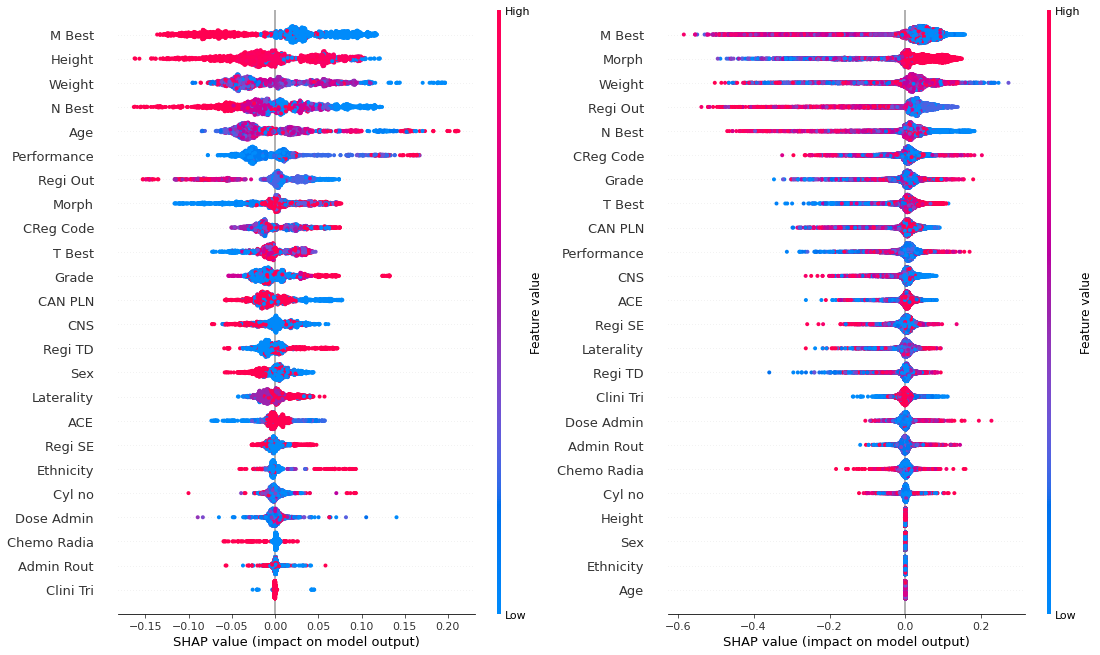}
         \caption{Global views of lung cancer cases in the Simulacrum (left: SHAP; right: CAFA). Uncontrollable features are: {\em Age}, {\em Ethnicity}, {\em Sex}, and {\em Height}.}
         \label{fig6a}
     \end{subfigure}
     \hfill
     \begin{subfigure}[b]{\linewidth}
         \centering
         \includegraphics[width=0.9\linewidth,trim={0cm 0.8cm 0cm 0.0cm},clip]{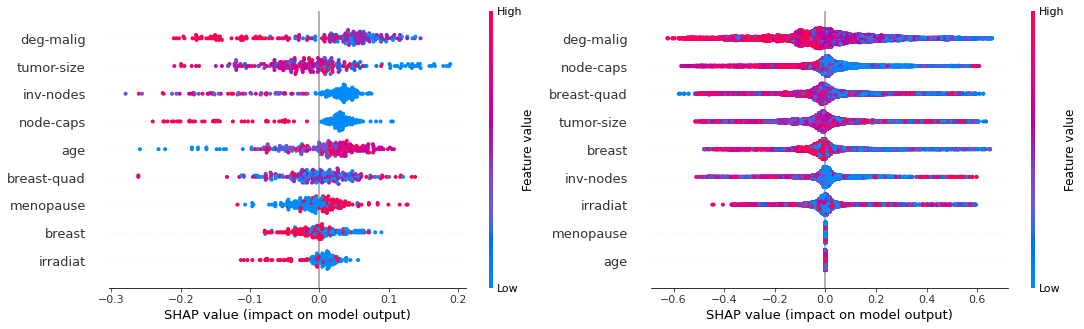}
         \caption{Global views of the UCI Breast Cancer dataset (left: SHAP; right: CAFA). Uncontrollable features are: {\em Age}, and {\em Menopause}.}
         \label{fig6b}
     \end{subfigure}
     \vspace{-20pt}
   \caption{Global explanations calculated using SHAP and CAFA on the Simulacurm Lung Cancer dataset and the Breast Cancer dataset. Same as Fig.~\ref{fig:lung_cancer_local}, we see that uncontrollable features in both datasets have importance 0; and CAFA produces similar results to SHAP for controllable features. 
   \label{fig:lung_breast_global}}
   \vspace{-20pt}
\end{figure}

We further study the influence of uncontrollable features with CAFA for global explanations. We randomly sample 100 instances from each dataset and compute global explanations with SHAP and CAFA. We produce ``violin plots'' using the summary plot function from the SHAP library. Fig.~\ref{fig:lung_breast_global} (a) and (b) illustrate global explanations for the lung and breast cancer datasets, respectively. There, the x-axis is the feature importance and the y-axis is the features. Color (red to blue) represents the value of a feature. 

For Fig.~\ref{fig:lung_breast_global} (a) and (b), the left-hand side figures show results from SHAP; and the right-hand side figures show results from CAFA. We can see that: (1) as seen in local explanation cases (Fig.~\ref{fig:lung_cancer_local}), all uncontrollable features are assigned the importance 0; (2) similar patterns on controllable features can be seen between SHAP and CAFA; (3) the orders of feature importance differ between SHAP and CAFA. We conclude that, for global explanations, CAFA precludes uncontrollable features from contributing to its explanations; and CAFA produces explanations similar to but distinct from SHAP explanations.

\vspace{-5pt}
\section{UK COVID-19 Case Study}
\label{sec:case}
\vspace{-5pt}
With the outbreak of the COVID-19 pandemic in December 2019, many countries have implemented some non-pharmaceutical control measures to contain the spread of the virus in the absence of effective prevention and treatment. In this case study, we use CAFA to study the effectiveness of non-pharmaceutical control measures implemented in the UK. 

We formulate the effectiveness of control measures as an XAI modelling problem. We focus on studying the relationship between control measures and the daily reproduction rate $R_{t}$. $R_{t}$ is one of the most important metrics used to measure the epidemic spread. A value greater than 1 suggests that the epidemic is expanding; a value less than 1 indicates that it is shrinking. We employ the approach presented in~\cite{Flaxman2020} for estimating $R_{t}$ from daily infection cases. 
We then pose the following classification problem:
\begin{quote}
\vspace{-5pt}
    {\em Given non-pharmaceutical control measures applied on a specific day, predict whether $R_t$ is smaller or greater than 1 on that day. }
    \vspace{-5pt}
\end{quote}
Having this prediction problem solved by a classifier, we use CAFA to identify control measures that make the most contribution to the prediction. Thus, by analysing the behaviour of the prediction model, we gain insight into the effectiveness of control measures.

\vspace{-15pt}
\subsubsection{Data Collection}
We have collected a dataset containing daily infection numbers and control measures from 04/01/2020 to 06/02/2021. Each instance consists of uncontrollable features (i.e., daily number of infections, cumulative cases, daily number of deaths and tests performed, temperature and humidity) and controllable features (i.e., implemented control measures). The numbers of daily cases, cumulative cases, deaths, and tests performed are collected from the Public Health England website\footnote{COVID-19 Dashboard (UK): \url{https://coronavirus.data.gov.uk}}. Control measure information is retrieved from Wikipedia\footnote{For example, for Wales the control measure data has been collected from\\ \url{https://en.wikipedia.org/wiki/Timeline\_of\_the\_COVID-19\_pandemic\_in\_Wales}
} and various news articles. 

We have considered control measures {\em school closures (SC)}, restrictions on {\em meeting friends and family indoors (MInd), meeting friends and family outdoors (MOut), domestic travel (DT), international travel (IT), hospitals and nursing home visits (HV), opening of cafes and restaurants (CR), accessing pubs and bars (PB), sports and leisure venues (SL)}, and {\em non-essential shops (NS)}. The values for control measures are binary, e.g, for ``school closure'', the values are ``open'' and ``closed''; for ``restrictions on meeting indoors'' the values are ``High'' (H) or ``Moderate'' (M). To consider the temporal effect of control measures, each feature is represented categorically. For instance, if they are open, then the ``school closure'' feature takes value 0; if the schools are closed for 0-5 days, then it takes value 1; etc. 

In total, we have collected 4,256 data points across 12 UK regions: East Midlands, East of England, London, North East, North West, South East, South West, West Midlands, Yorkshire and Humber, Northern Ireland, Scotland and Wales. To remove noise and achieve a more accurate $R_{t}$ estimation, we drop data points with cumulative cases less than 20 for each region and keep 3,936 instances. A sliding-window mean filter of size 3 has been used to filter noise in daily cases.

\vspace{-15pt}
\subsubsection{Experimental Results}
\begin{figure}
\centerline\centering{\includegraphics[width=0.9\textwidth,trim={0cm 0.8cm 0cm 0.0cm},clip]{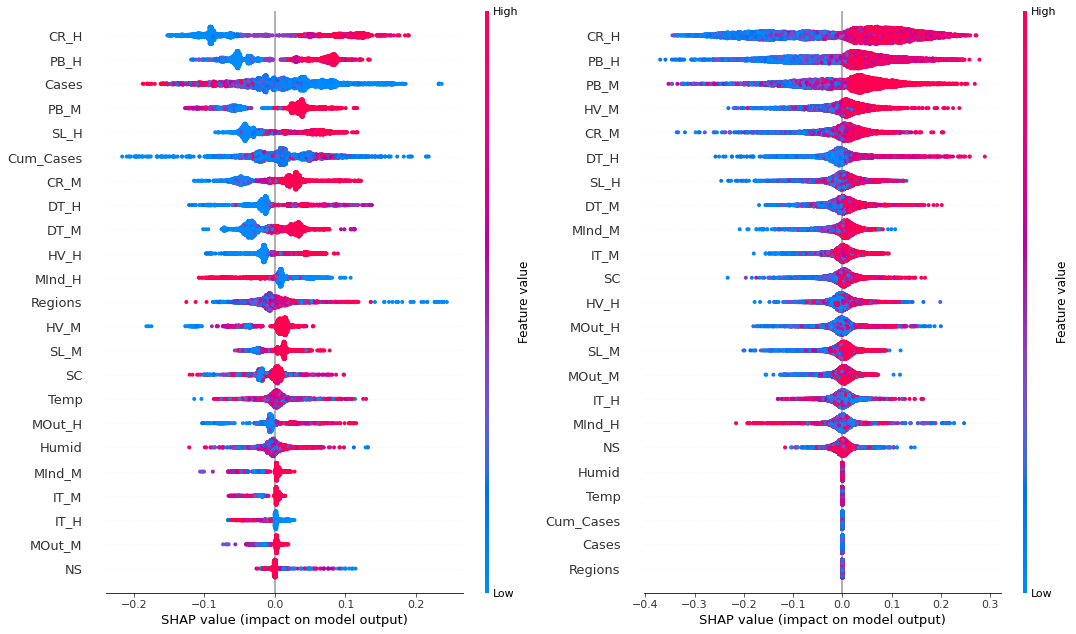}}
\caption{Global views of the COVID dataset (SHAP Left; CAFA Right). Uncontrollable features are: {\em Humidity (Humid)}, {\em Temperature (Temp)}, {\em Cumulative Cases (Cum\_cases)}, {\em Daily Infections (Cases)} and {\em Regions}.
\label{fig4}}
\vspace{-20pt}
\end{figure}

We split the dataset as 70\% for training and 30\% for testing, and
use a random forest classifier. We achieve a high prediction accuracy of 94.4\%. 
Since we aim to obtain a bird's-eye view of how control measures are affecting the disease, we focus on calculating global explanations. To this end, for each instance $\mbf{x}$, we generate $D_\mbf{x}$ with $K=500$. $\pi_x$ is the average distance between any two instances; $\omega_i=1$. By following Algorithm~\ref{alg:CAFA}, we obtain feature importance with CAFA. The global explanations are shown in Fig.~\ref{fig4}, right-hand side, with SHAP results shown on the left. 

SHAP results presented on the left shows that the number of daily cases and cumulative cases both have strong impact in predicting $R_t$. However, as both are uncontrollable, knowing that they have strong influence to the prediction does not help us understand the effectiveness of control measures. With CAFA (Fig.~\ref{fig4} right-hand side), the importance of all uncontrollable features are assigned to 0. Overall, we observe that:
\begin{itemize}[nolistsep]
\item
SHAP considers {\em High Restriction on Cafes and Restaurants Access (CR\_H)}, {\em High Restriction on Pubs and Bars Access (PB\_H)}, {\em Number of Daily Infections (Cases)}, {\em Number of Daily Infections (Cases)}, {\em Medium Restriction on Pubs and Bars Access (PB\_M)}, and {\em High Restriction Sport and Leisure Facilities (SL\_H)} as the top five effective control measures; whereas 

\item
CAFA considers {\em CR\_H, PB\_H, PB\_M}, {\em Medium Restriction on Hospital and Nursing Home Visits (HV\_M)} and {\em Medium Restriction on Cafes and Restaurants Access (CR\_M)} as the top five effective control measures.
\end{itemize}

\noindent
CAFA's results are in alignment with WHO's COVID-19 guideline\footnote{Coronavirus disease (COVID-19): How is it transmitted?:\\ 
\sloppy \burl{https://www.who.int/news-room/q-a-detail/coronavirus-disease-covid-19-how-is-it-transmitted}} stating the ``Three C's'' rule that the virus is more transmissible with (1) {\em Crowded places}; (2) {\em Close-contact settings}; and (3) {\em Confined and enclosed spaces with poor ventilation}.Focusing on restricting access to cafes and restaurants as well as pubs and bars seem to be a very reasonable strategy in reducing the virus transmission, for the reason that these are the most prominent locations meeting the Three C's for most of the population.

\vspace{-10pt}
\section{Related Work}
\label{sec:related}
\vspace{-5pt}
There has been some research conducted to extend feature attribution algorithms to achieve more meaningful explanations. For example, \cite{Aas2021} extended the Kernal SHAP method to handle dependent features through different approaches to estimate the conditional distribution. Experiments over simulated datasets suggest that the dependencies between features are handled properly using proposed Shapley value approximations. An aggregation of the Shapley values of dependent features was also introduced to ease the interpretation and use of the Shapley values. In \cite{Ribeiro2018}, a model-agnostic explanation approach `anchors' was proposed based on if-then rules, which depends on input perturbation to approximate local explanations. Experimental results over classification,
structured prediction, and text generation machine learning tasks demonstrated the usefulness of anchors. In \cite{agarwal21c}, a variant of LIME for continuous data was proposed. Theoretical analysis was performed to derive explicit closed form expressions for the explanations output. It was also demonstrated that
post hoc explanation methods will converge to the same explanations when the number of perturbed samples used by these methods is large.

Great effort has been put into studying the effectiveness of control measures for containing the COVID-19 pandemic. For example, in~\cite{leung2020first}, the authors estimated the instantaneous reproduction number ($R_{t}$) of COVID-19 in four Chinese cities and ten provinces. They found that though aggressive non-pharmaceutical interventions (e.g., city lockdown) had abated the first wave of COVID-19 outside of Hubei. The effect of physical distancing measures on the progression of the COVID-19 epidemic was explored in \cite{prem2020effect}. An extensive simulation based on an age-structured susceptible-exposed-infected-removed model \cite{klepac2011stage} was carried out. The simulation results show that sustained physical distancing measures have a potential to reduce the magnitude of the epidemic peak of COVID-19. The impact of physical distancing measures in the UK was evaluated through comparing the contact patterns during the ``lockdown'' to patterns of social contact made before the epidemic \cite{jarvis2020quantifying}. It was found that the estimated change in reproduction number significantly decreased, suggesting that the physical distancing measures adopted by the UK public would probably lead to a decline in cases.

\vspace{-10pt}
\section{Conclusion}
\label{sec:cln}
\vspace{-5pt}

Feature attribution XAI algorithms tell users the relative contribution of a feature in a prediction, which can help users gain insight by shedding light onto the underlying patterns in large datasets. However, existing feature attribution algorithms treat controllable and uncontrollable features homogeneously, which may lead to incorrect estimation of the importance of controllable features. 
In this paper, we proposed CAFA to compute the relative importance of controllable factors through generating perturbed instances. Specifically, for each prediction instance, CAFA creates a dataset by selectively perturbing features representing controllable factors while leaving uncontrollable ones unchanged and then computing the global explanation on the generated dataset as the local explanation for the prediction instance. 

We tested CAFA on two existing medical datasets, lung cancer data from the Simulacrum dataset and the UCI breast cancer dataset. Experimental results show that with CAFA, although the prediction model is built over all features, the explanations of controllable features are not interfered with by the uncontrollable ones. We further applied CAFA in a case study on understanding the effectiveness of COVID-19 non-pharmaceutical control measures implemented in the UK during the period of January 2020 to February 2021. We found that restricting access to cafes and restaurants as well as pubs and bars are the most effective measures in containing the disease, represented by reaching an $R_t$ value smaller than 1.

\vspace{-10pt}
\bibliographystyle{splncs04}
\bibliography{Ref}

\begin{thebibliography}{10}
\providecommand{\url}[1]{\texttt{#1}}
\providecommand{\urlprefix}{URL }
\providecommand{\doi}[1]{https://doi.org/#1}

\bibitem{Aas2021}
Aas, K., et~al.: Explaining individual predictions when features are dependent:
  More accurate approximations to shapley values. AI Journal  \textbf{298}
  (2021)

\bibitem{Adadi2018}
Adadi, A., Berrada, M.: Peeking inside the black-box: A survey on explainable
  artificial intelligence (xai). IEEE Access  \textbf{6} (2018)

\bibitem{agarwal21c}
Agarwal, S., et~al.: Towards the unification and robustness of perturbation and
  gradient based explanations. In: Proc. of ICML. vol.~139, pp. 110--119 (2021)

\bibitem{Antoniadi2021}
Antoniadi, A.M., et~al.: Current challenges and future opportunities for xai in
  machine learning-based clinical decision support systems: A systematic
  review. Applied Sciences (Switzerland)  \textbf{11} (2021)

\bibitem{Draper06}
Draper, L.: Breast cancer: Trends, risks, treatments, and effects. AAOHN
  Journal  \textbf{54}(10),  445--453 (2006)

\bibitem{Duell21}
Duell, J., et~al.: A comparison of explanations given by explainable artificial
  intelligence methods on analysing electronic health records. In: Proc. of BHI
  (2021)

\bibitem{Fan20b}
Fan, X., et~al.: An investigation of covid-19 spreading factors with
  explainable ai techniques. International Journal of Information Technology
  \textbf{26} (2020)

\bibitem{Flaxman2020}
Flaxman, S., et~al.: {Estimating the effects of non-pharmaceutical
  interventions on COVID-19 in Europe}. Nature  (2020)

\bibitem{Honegger18}
Honegger, M.: Shedding light on black box machine learning algorithms:
  Development of an axiomatic framework to assess the quality of methods that
  explain individual predictions. CoRR  \textbf{abs/1808.05054} (2018)

\bibitem{jarvis2020quantifying}
Jarvis, C.I., et~al.: Quantifying the impact of physical distance measures on
  the transmission of covid-19 in the uk. BMC medicine  \textbf{18},  1--10
  (2020)

\bibitem{klepac2011stage}
Klepac, P., Caswell, H.: The stage-structured epidemic: linking disease and
  demography with a multi-state matrix approach model. T. Ecology
  \textbf{4}(3), ~301 (2011)

\bibitem{leung2020first}
Leung, K., et~al.: First-wave covid-19 transmissibility and severity in china
  outside hubei after control measures, and second-wave scenario planning: a
  modelling impact assessment. The Lancet  (2020)

\bibitem{Liu21}
Liu, S., et~al.: An investigation of the impact of covid-19 non-pharmaceutical
  interventions and economic support policies on foreign exchange markets with
  explainable ai techniques. In: Proc. of XAI-FIN21 (2021)

\bibitem{Lundberg2017}
Lundberg, S.M., Lee, S.I.: {A unified approach to interpreting model
  predictions}. In: Proc. of NIPS (2017)

\bibitem{Lundberg19}
Lundberg, S.M., et~al.: Explainable {AI} for trees: From local explanations to
  global understanding. CoRR  \textbf{abs/1905.04610} (2019)

\bibitem{molnar2019}
Molnar, C.: Interpretable Machine Learning, A Guide for Making Black Box Models
  Explainable (2019), \url{https://christophm.github.io/interpretable-ml-book/}

\bibitem{prem2020effect}
Perm, K., et~al.: The effect of control strategies to reduce social mixing on
  outcomes of the covid-19 epidemic in wuhan, china: a modelling study. The
  Lancet Public Health  (2020)

\bibitem{Ribeiro16}
Ribeiro, M.T., Singh, S., Guestrin, C.: "{W}hy should {I} trust you?":
  Explaining the predictions of any classifier. In: Proc. of SIGKDD (2016)

\bibitem{Ribeiro2018}
Ribeiro, M.T., Singh, S., Guestrin, C.: Anchors: High-precision model-agnostic
  explanations. In: Proc. of AAAI (2018)

\bibitem{Slack20}
Slack, D., et~al.: Fooling {LIME} and {SHAP:} adversarial attacks on post hoc
  explanation methods. In: Proc. of AIES. pp. 180--186. {ACM} (2020)

\bibitem{Tjoa2020}
Tjoa, E., Guan, C.: A survey on explainable artificial intelligence (xai):
  Toward medical xai. IEEE Trans. on Neural Networks and Learning Systems
  (2020)

\bibitem{Yalcin21}
Yalcin, O., Fan, X., Liu, S.: Evaluating the correctness of explainable {AI}
  algorithms for classification. CoRR  \textbf{abs/2105.09740} (2021)

\bibitem{IndreZ2016}
Žliobaitė, I., Custers, B.: Using sensitive personal data may be necessary
  for avoiding discrimination in data-driven decision models. AI and Law
  \textbf{24} (2016)

\end{thebibliography}

\end{document}